\documentclass{article} % For LaTeX2e
\usepackage{nips15submit_e,times}
\usepackage{hyperref}
\usepackage{url}
\pdfoutput=1 %for posting to arXiv

\usepackage{graphicx}
\usepackage{subfigure}
\usepackage{caption}
\usepackage{booktabs}

\usepackage{tabularx}
\usepackage{microtype}

\title{Learning both Weights and Connections for Efficient Neural Networks}

\author{
Song Han \\
Stanford University \\
\texttt{songhan@stanford.edu} \\
\And
Jeff Pool \\
NVIDIA \\
\texttt{jpool@nvidia.com} \\
\AND
John Tran \\
NVIDIA \\
\texttt{johntran@nvidia.com} \\
\And
William J. Dally \\
Stanford University \\
NVIDIA \\
\texttt{dally@stanford.edu}
}

\nipsfinalcopy % Uncomment for camera-ready version

\begin{document}

\maketitle

\begin{abstract}
Neural networks are both computationally intensive and memory intensive, making them difficult to deploy on embedded systems. Also, conventional networks fix the architecture before training starts; as a result, training cannot improve the architecture. To address these limitations, we describe a method to reduce the storage and computation required by neural networks by an order of magnitude without affecting their accuracy by learning only the important connections. Our method prunes redundant connections using a three-step method. First, we train the network to learn which connections are important. Next, we prune the unimportant connections. Finally, we retrain the network to fine tune the weights of the remaining connections. On the ImageNet dataset, our method reduced the number of parameters of AlexNet by a factor of $\bf{9\times}$, from 61 million to 6.7 million, without incurring accuracy loss. Similar experiments with VGG-16 found that the total number of parameters can be reduced by $\bf{13\times}$, from 138 million to 10.3 million, again with no loss of accuracy. 

\end{abstract}

\section{Introduction}

Neural networks have become ubiquitous in applications ranging 
from computer vision \cite{hinton12} to speech recognition \cite{graves2005framewise} and natural language processing \cite{NLP_scratch}. We consider convolutional neural networks used for computer vision tasks which have grown over time. 
In 1998 Lecun \textit{et al.} designed a CNN model LeNet-5 with less than 1M parameters to classify handwritten digits \cite{lecun1998gradient}, while in 2012, Krizhevsky \textit{et al.} \cite{hinton12} won the ImageNet competition with 60M parameters. 
Deepface classified human faces with 120M parameters \cite{deepface}, and Coates \textit{et al.} \cite{cots} scaled up a network to 10B parameters.

\begin{figure}[htb]
  \centering
  \begin{minipage}[c]{0.65\textwidth}
    \centering

\begin{tabular}{@{}lll@{}}
\toprule
Operation            & Energy {[}pJ{]} & Relative Cost \\ \midrule
32 bit int ADD       & 0.1             & 1             \\
32 bit float ADD        & 0.9             & 9           \\
32 bit Register File & 1               & 10          \\
32 bit int MULT      & 3.1             & 31          \\
32 bit float MULT       & 3.7             & 37          \\
32 bit SRAM Cache    & 5               & 50          \\
\bf{32 bit DRAM Memory}  & \bf{640}        & \bf{6400}   \\ \bottomrule
\end{tabular}

  \end{minipage}
  \begin{minipage}[c]{0.32\textwidth}
    \scalebox{1}[1] {\includegraphics[width=1.05\textwidth]{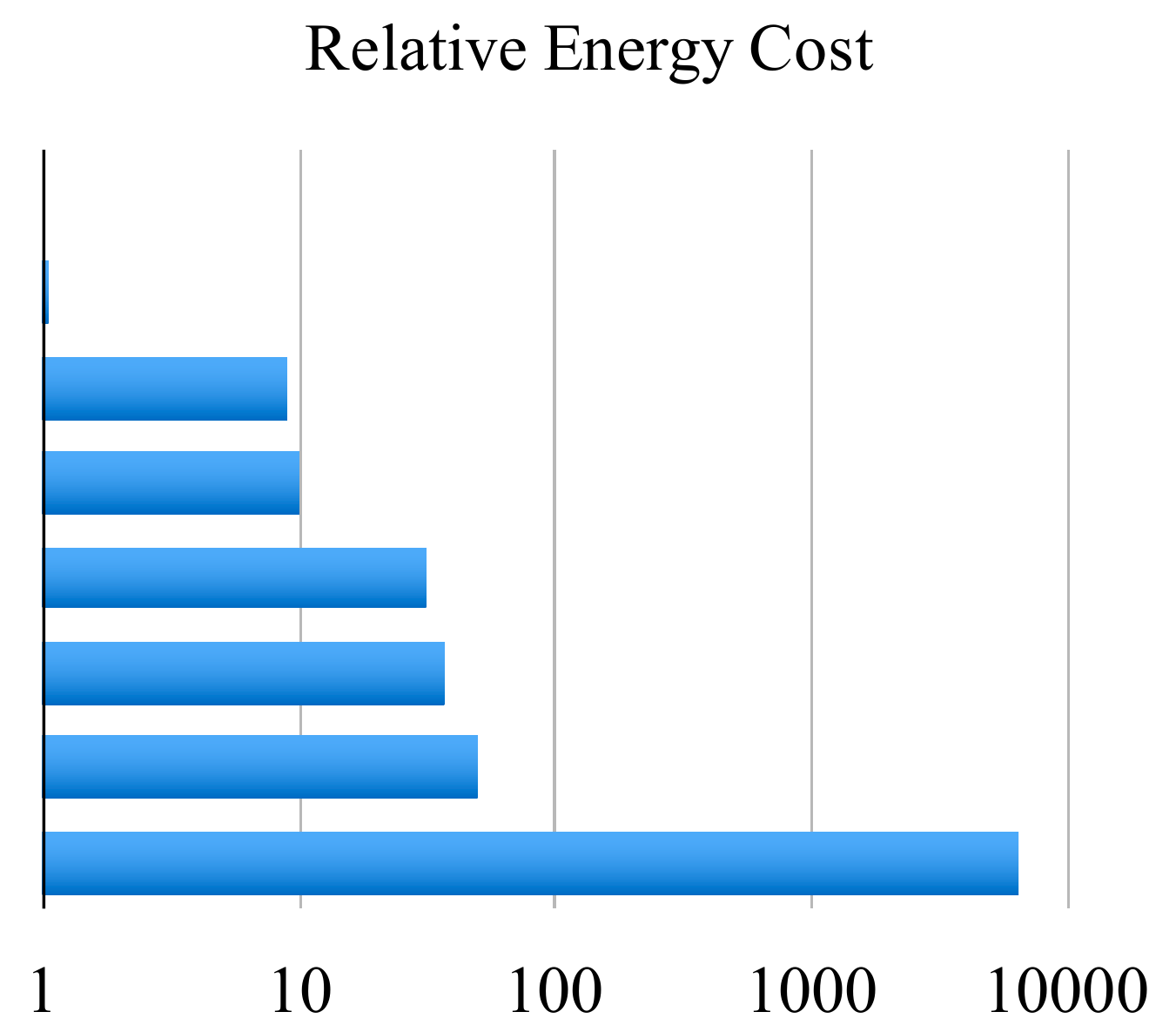}}
  \end{minipage}
  \caption{Energy table for 45nm CMOS process \cite{markenergy}. Memory access is 3 orders of magnitude more energy expensive than simple arithmetic.}
  \label{energy}
\end{figure}

While these large neural networks are very powerful, their size consumes considerable storage, memory bandwidth, and computational resources.
For embedded mobile applications, these resource demands become prohibitive.
Figure \ref{energy} shows the energy cost of basic arithmetic and memory operations in a 45nm CMOS process.
From this data we see the energy per connection is dominated by 
memory access and ranges from 5pJ for 32 bit coefficients in on-chip SRAM to 640pJ for 32bit coefficients in off-chip DRAM \cite{markenergy}.
Large networks do not fit in on-chip storage and hence require the more costly DRAM accesses.
Running a 1 billion connection neural network, for example, at 20Hz would require $(20Hz)(1G)(640pJ)=12.8W$ just for DRAM access - well beyond the power envelope of a typical mobile device.
Our goal in pruning networks is to reduce the energy required to run such
large networks so they can run in real time on mobile devices. The model size reduction from pruning also facilitates storage and
transmission of mobile applications incorporating DNNs.

To achieve this goal, we present a method to prune network connections in a manner that preserves the original accuracy.  
After an initial training phase, we remove all connections whose weight is lower than a threshold.
This pruning converts a dense, fully-connected layer to a sparse layer.
This first phase learns the topology of the networks --- learning which connections are important and removing the unimportant connections.
We then retrain the sparse network so the remaining connections can 
compensate for the connections that have been removed.
The phases of pruning and retraining may be repeated iteratively
to further reduce network complexity.
In effect, this training process learns the network connectivity in addition to the weights - much as in the mammalian brain \cite{catlearningconnections}\cite{nature}, where synapses are created in the first few months of a child's development, followed by gradual “pruning” of little-used connections, falling to typical adult values.

\section{Related Work}

Neural networks are typically over-parameterized, and there is significant redundancy for deep learning models \cite{denil2013predicting}. This results in a waste of both computation and memory. There have been various proposals to remove the redundancy: Vanhoucke \textit{et al.} \cite{vincentGoogle2011improving} explored a fixed-point implementation with 8-bit integer (vs 32-bit floating point) activations. Denton \textit{et al.} \cite{denton2014LinearStructure} exploited the linear structure of the neural network by finding an appropriate low-rank approximation of the parameters and keeping the accuracy within 1\% of the original model. With similar accuracy loss, Gong \textit{et al.} \cite{gong2014compressing} compressed deep convnets using vector quantization. These approximation and quantization techniques are orthogonal to network pruning, and they can be used together to obtain further gains \cite{han2015deep}. 

There have been other attempts to reduce the number of parameters of neural networks by replacing the fully connected layer with global average pooling. The Network in Network architecture \cite{NiN} and GoogLenet \cite{szegedy2014GoogLenet} achieves state-of-the-art results on several benchmarks by adopting this idea. However, transfer learning, i.e. reusing features learned on the ImageNet dataset and applying them to new tasks by only fine-tuning the fully connected layers, is more difficult with this approach. This problem is noted by Szegedy \textit{et al.} \cite{szegedy2014GoogLenet} and motivates them to add a linear layer on the top of their networks to enable transfer learning. 

Network pruning has been used both to reduce network complexity
and to reduce over-fitting.
An early approach to pruning was biased weight decay \cite{hanson1989comparing}.
Optimal Brain Damage \cite{Cun90optimalbrain} and Optimal Brain Surgeon \cite{hassibi1993second} prune networks to reduce the number of connections based on the Hessian of the loss function and suggest that such pruning is more accurate than magnitude-based pruning such as weight decay. However, second order derivative needs additional computation.

HashedNets \cite{chen2015compressing} is a recent technique to reduce model sizes by using a hash function to randomly group connection weights into hash buckets, so that all connections within the same hash bucket share a single parameter value. This technique may benefit from pruning. As pointed out in Shi \textit{et al.} \cite{shi2009hash} and Weinberger \textit{et al.} \cite{weinberger2009feature}, sparsity will minimize hash collision making feature hashing even more effective. HashedNets may be used together with pruning to give even better parameter savings.

\section{Learning Connections in Addition to Weights}

\begin{figure}[t]
\centering
\noindent\begin{minipage}{.5\linewidth}
\centering
\includegraphics[scale=0.41]{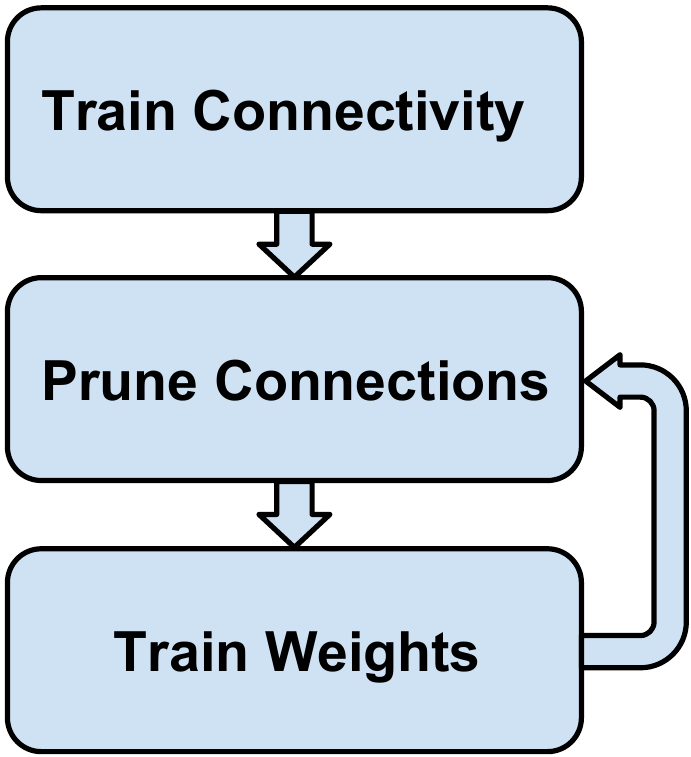}
\caption{Three-Step Training Pipeline.}
\label{fig:pip}
\end{minipage}%
\begin{minipage}{.5\linewidth}
\centering
\includegraphics[scale=0.35]{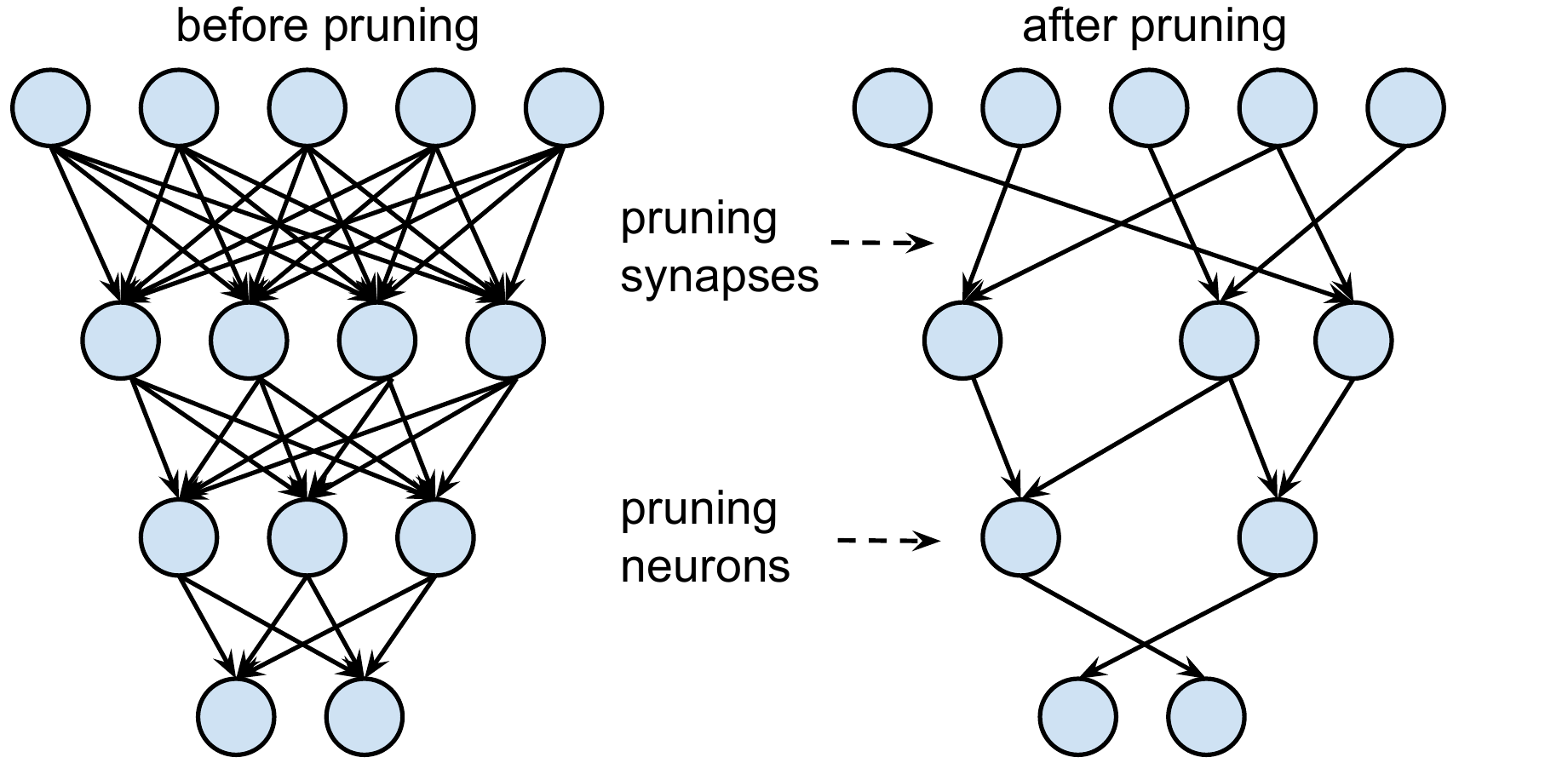}
\caption{Synapses and neurons before and after pruning.}
\label{fig:network}
\end{minipage}
\end{figure}

Our pruning method employs a three-step process, as illustrated in Figure \ref{fig:pip},
which begins by learning the connectivity via normal network training.
Unlike conventional training, however, we are not learning the final values of the weights, but rather we are learning which connections are important.

The second step is to prune the low-weight connections.
All connections with weights below a threshold are removed from the network --- converting a dense network into a sparse network, as shown in Figure \ref{fig:network}. 
The final step retrains the network to learn the final weights for the remaining sparse connections.
This step is critical. If the pruned network is used without retraining, accuracy is significantly impacted.

\subsection{Regularization}
Choosing the correct regularization impacts the performance of pruning and retraining. 
L1 regularization penalizes non-zero parameters resulting in more parameters near zero.
This gives better accuracy after pruning, but before retraining.
However, the remaining connections are not as good as with L2 regularization, resulting in lower accuracy after retraining. Overall, L2 regularization gives the best pruning results. This is further discussed in experiment section.

\subsection{Dropout Ratio Adjustment}
Dropout \cite{dropout} is widely used to prevent over-fitting, and this also applies to retraining. 
During retraining, however, the dropout ratio must be adjusted to account for the change in model capacity. In dropout, each parameter is probabilistically dropped during training, but will come back during inference. 
In pruning, parameters are dropped forever after pruning and have no chance to come back during both training and inference. As the parameters get sparse, the classifier will select the most informative predictors and thus have much less prediction variance, which reduces over-fitting. As pruning already reduced model capacity, the retraining dropout ratio should be smaller. 

Quantitatively, let $C_i$ be the number of connections in layer $i$, $C_{io}$ for the original network, $C_{ir}$ for the network after retraining, $N_i$ be the  number of neurons in layer i. Since dropout works on neurons, and $C_i$ varies quadratically with $N_i$, according to Equation \ref{eq_param} thus the dropout ratio after pruning the parameters should follow Equation \ref{eq_dropout}, where $D_o$ represent the original dropout rate, $D_r$ represent the dropout rate during retraining.

\noindent\begin{minipage}{.5\linewidth}
\begin{equation}
C_i = N_i N_{i-1}
\label{eq_param}
\end{equation}
\end{minipage}%
\begin{minipage}{.5\linewidth}
\begin{equation}
D_r = D_o \sqrt{\frac{C_{ir}}{C_{io}}}
\label{eq_dropout}
\end{equation}
\end{minipage}

\subsection{Local Pruning and Parameter Co-adaptation} 
During retraining, it is better to retain the weights from
the initial training phase for the connections that survived 
pruning than it is to re-initialize the pruned layers. 
CNNs contain fragile co-adapted features \cite{yosinski2014transferable}: gradient descent is able to find a good solution when the network is initially trained, but not after re-initializing some layers and retraining them. 
So when we retrain the pruned layers, we should keep the surviving parameters instead of re-initializing them.

Retraining the pruned layers starting with retained weights requires less computation because we don't have to back propagate through the entire network. Also, neural networks are prone to suffer the vanishing gradient problem \cite{bengio1994vanish} as the networks get deeper, which makes pruning errors harder to recover for deep networks. To prevent this, we fix the parameters for CONV layers and only retrain the FC layers after pruning the FC layers, and vice versa.

\subsection{Iterative Pruning} Learning the right connections is an iterative process. Pruning followed by a retraining is one iteration, after many such iterations the minimum number connections could be found. Without loss of accuracy, this method can boost pruning rate from $5\times$ to $9\times$ on AlexNet compared with single-step aggressive pruning. Each iteration is a greedy search in that we find the best connections. We also experimented with probabilistically pruning parameters based on their absolute value, but this gave worse results.

\subsection{Pruning Neurons} After pruning connections, neurons with zero input connections or zero output connections may be safely pruned.
This pruning is furthered by removing all connections to or from a pruned neuron.
The retraining phase automatically arrives at the result where dead neurons will have both zero input connections and zero output connections. 
This occurs due to gradient descent and regularization. 
A neuron that has zero input connections (or zero output connections) will have no contribution to the final loss, leading the gradient to be zero for its output connection (or input connection), respectively. Only the regularization term will push the weights to zero. Thus, the dead neurons will be automatically removed during retraining.

\section{Experiments}
We implemented network pruning in Caffe \cite{jia2014caffe}. 
Caffe was modified to add a mask which 
disregards pruned parameters during network operation for each weight tensor.
The pruning threshold is chosen as a quality parameter multiplied
by the standard deviation of a layer's weights. We carried out the experiments on Nvidia TitanX and GTX980 GPUs.

We pruned four representative networks: Lenet-300-100 and Lenet-5 on MNIST, together with AlexNet and VGG-16 on ImageNet.
The network parameters and accuracy \footnote{Reference model is from Caffe model zoo, accuracy is measured without data augmentation} before and after pruning are shown in Table \ref{table:results}.

\begin{table}[t]
\centering
\caption{Network pruning can save $9\times$ to $13\times$ parameters with no drop in predictive performance.}
\begin{tabular}{lllll}
\multicolumn{1}{l|}{Network}              & Top-1 Error& \multicolumn{1}{l|}{Top-5 Error} & \multicolumn{1}{l|}{Parameters}  & \begin{tabular}[c]{@{}l@{}}Compression\\       Rate\end{tabular}  \\ \hline
\multicolumn{1}{l|}{LeNet-300-100 Ref}    & 1.64\%      & \multicolumn{1}{l|}{-}           & \multicolumn{1}{l|}{267K}       &        \\
\multicolumn{1}{l|}{LeNet-300-100 Pruned} & 1.59\%      & \multicolumn{1}{l|}{-}           & \multicolumn{1}{l|}{\bf{22K}}   & $\bf{12\times}$ \\ \hline
\multicolumn{1}{l|}{LeNet-5 Ref}          & 0.80\%      & \multicolumn{1}{l|}{-}           & \multicolumn{1}{l|}{431K}       &        \\
\multicolumn{1}{l|}{LeNet-5 Pruned}       & 0.77\%      & \multicolumn{1}{l|}{-}           & \multicolumn{1}{l|}{\bf{36K}}   & $\bf{12\times}$ \\ \hline
\multicolumn{1}{l|}{AlexNet Ref}          & 42.78\%     & \multicolumn{1}{l|}{19.73\%}     & \multicolumn{1}{l|}{61M}        &        \\
\multicolumn{1}{l|}{AlexNet Pruned}       & 42.77\%     & \multicolumn{1}{l|}{19.67\%}     & \multicolumn{1}{l|}{\bf{6.7M}}      & 
$\bf{9\times}$ \\ \hline
\multicolumn{1}{l|}{VGG-16 Ref}           & 31.50\%     & \multicolumn{1}{l|}{11.32\%}      & \multicolumn{1}{l|}{138M}        &        \\
\multicolumn{1}{l|}{VGG-16 Pruned}        & 31.34\%     & \multicolumn{1}{l|}{10.88\%}      & \multicolumn{1}{l|}{\bf{10.3M}}      & 
$\bf{13\times}$     \\
\end{tabular}
\label{table:results}
\end{table}

\subsection{LeNet on MNIST}

We first experimented on MNIST dataset with the LeNet-300-100 and LeNet-5 networks \cite{lecun1998gradient}. LeNet-300-100 is a fully connected network with two hidden layers, with 300 and 100 neurons each, which achieves 1.6\% error rate on MNIST. LeNet-5 is a convolutional network that has two convolutional layers and two fully connected layers, which  achieves 0.8\% error rate on MNIST. After pruning, the network is retrained with $1/10$ of the original network's original learning rate.  Table \ref{table:results} shows pruning saves $12\times$ parameters on these networks.  For each layer of the network the table shows (left to right) the original number of weights, the number of floating point operations to compute that layer's activations, the average percentage of activations that are non-zero, the percentage of non-zero weights after pruning, and the percentage of actually required floating point operations.

\begin{table}[t]
\centering
\caption{For Lenet-300-100, pruning reduces the number of weights by $12\times$ and computation by $12\times$.}
\begin{tabular}{l|lllll}
Layer & Weights    & FLOP  & Act\%     & Weights\%  & FLOP\%  \\ \hline
fc1   & 235K       & 470K   & 38\%      & 8\%        & 8\%             \\
fc2   & 30K        & 60K    & 65\%      & 9\%       &  4\%           \\
fc3   & 1K         & 2K     & 100\%     & 26\%       & 17\%           \\ \hline
Total & 266K       & 532K   & 46\%    & \bf{8\%}   & \bf{8\%}            
\end{tabular}
\end{table}

\begin{table}[t!]
\centering
\caption{For Lenet-5, pruning reduces the number of weights by $12\times$ and computation by $6\times$.}
\begin{tabular}{l|lllll}
Layer & Weights   & FLOP   & Act\%     & Weights\%    & FLOP\%     \\ \hline
conv1 & 0.5K      & 576K   & 82\%      & 66\%         & 66\%              \\
conv2 & 25K       & 3200K  & 72\%      & 12\%         & 10\%              \\ \hline
fc1   & 400K      & 800K   & 55\%      & 8\%         & 6\%               \\
fc2   & 5K        & 10K    & 100\%     & 19\%         & 10\%              \\ \hline
Total & 431K      & 4586K  & 77\% & \bf{8\%}    &\bf{16\%}         
\end{tabular}
\end{table}

\begin{figure}[t!]
\centering
\scalebox{1}[0.8]{\includegraphics[scale=2.0]{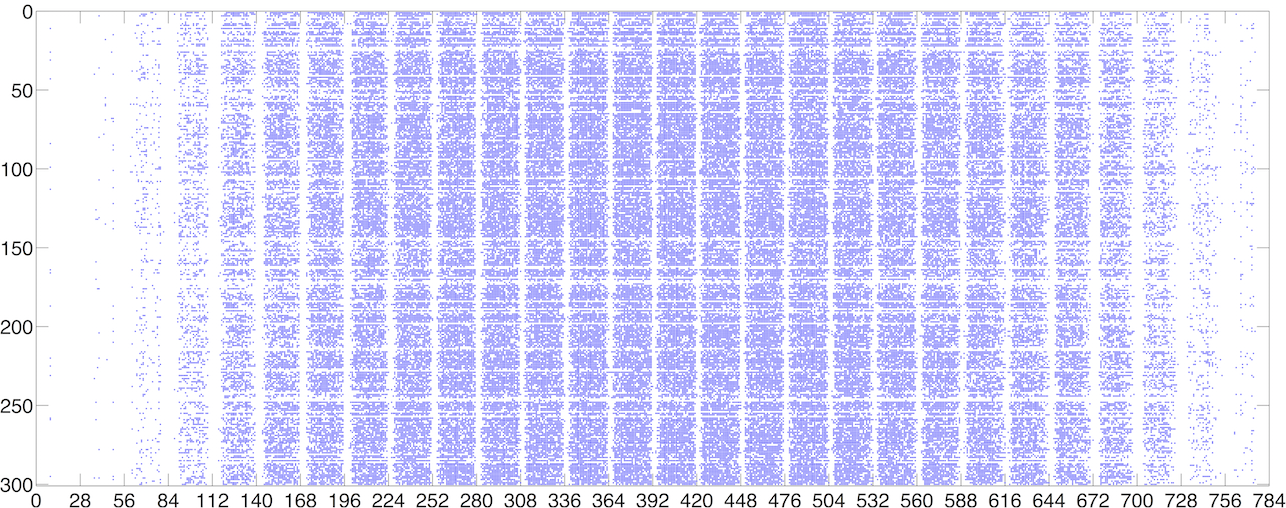}}
\caption{Visualization of the first FC layer's sparsity pattern of Lenet-300-100. It has a banded structure repeated 28 times, which correspond to the un-pruned parameters in the center of the images, since the digits are written in the center.}
\label{fig:region}
\end{figure}

An interesting byproduct is that network pruning detects visual attention regions. Figure \ref{fig:region} shows the sparsity pattern of the first fully connected layer of LeNet-300-100, the matrix size is $784 * 300$. It has 28 bands, each band's width 28, corresponding to the $28 \times 28$ input pixels. The colored regions of the figure, indicating non-zero parameters, correspond to the center of the image. 
Because digits are written in the center of the image, these are the important parameters. The graph is sparse on the left and right, corresponding to the less important regions on the top and bottom of the image.  After pruning, the neural network finds the center of the image more important, and the connections to the peripheral regions are more heavily pruned.

\subsection{AlexNet on ImageNet}

We further examine the performance of pruning on the ImageNet ILSVRC-2012 dataset, which has 1.2M training examples and 50k validation examples. We use the AlexNet Caffe model as the reference model, which has 61 million parameters across 5 convolutional layers and 3 fully connected layers. The AlexNet Caffe model achieved a top-1 accuracy of 57.2\% and a top-5 accuracy of 80.3\%. The original AlexNet took 75 hours to train on NVIDIA Titan X GPU. After pruning, the whole network is retrained with $1/100$ of the original network's initial learning rate. It took 173 hours to retrain the pruned AlexNet.  Pruning is not used when iteratively prototyping the model, but rather used for model reduction when the model is ready for deployment.  Thus, the retraining time is less a concern. Table \ref{table:results} shows that AlexNet can be pruned to $1/9$ of its original size without impacting accuracy, and the amount of computation can be reduced by $3\times$.

\begin{table}[t]
\caption{For AlexNet, pruning reduces the number of weights by $9\times$ and computation by $3\times$.}
\label{table:stats}
  \begin{minipage}[c]{0.62\textwidth}
\begin{tabularx}{\textwidth}{p{0.8cm}|p{1.1cm}p{0.8cm}p{1cm}p{1.4cm}p{1cm}}
Layer & Weights    & FLOP  & Act\%    & Weights\%   & FLOP\% \\ \hline
conv1 & 35K        & 211M  & 88\%     & 84\%        & 84\%   \\
conv2 & 307K       & 448M  & 52\%     & 38\%        & 33\%   \\
conv3 & 885K       & 299M  & 37\%     & 35\%        & 18\%   \\
conv4 & 663K       & 224M  & 40\%     & 37\%        & 14\%   \\
conv5 & 442K       & 150M  & 34\%     & 37\%        & 14\%   \\ \hline
fc1   & 38M        & 75M   & 36\%     & 9\%         & 3\%    \\
fc2   & 17M        & 34M   & 40\%     & 9\%         & 3\%    \\
fc3   & 4M         & 8M    & 100\%    & 25\%        & 10\%   \\ \hline
Total & 61M        & 1.5B  & 54\%     & \bf{11\%}   & \bf{30\%}  
\end{tabularx}
  \end{minipage}
  \begin{minipage}[c]{0.37\textwidth}
    \includegraphics[width=1\textwidth]{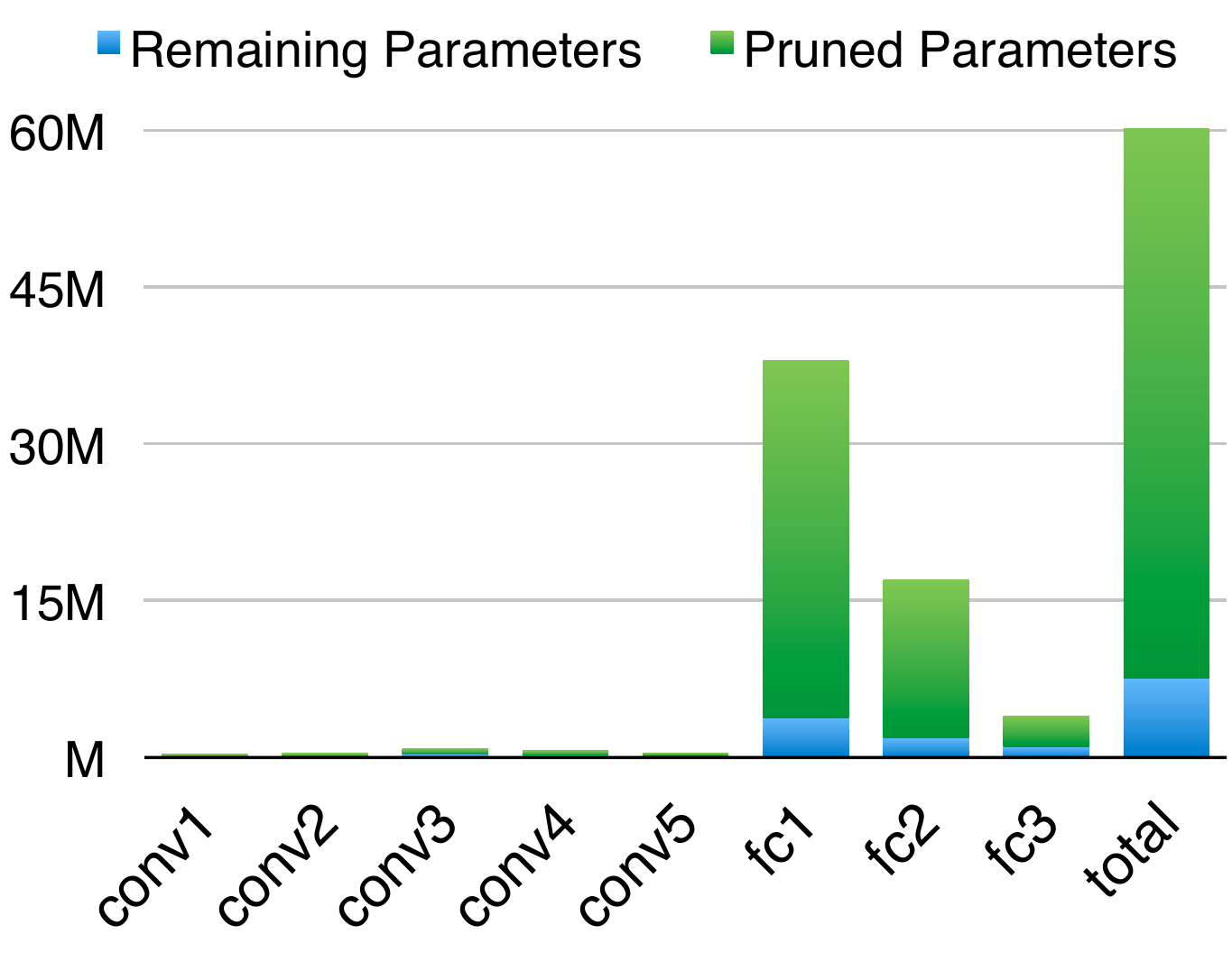}
  \end{minipage}

\end{table}

\begin{table}[t]
\centering
\caption{For VGG-16, pruning reduces the number of weights by $12\times$ and computation by $5\times$.}
\label{table:vggresults}
\begin{tabular}{l|lllll}
Layer    & Weights & FLOP  & Act\% & Weights\% & FLOP\% \\ \hline
conv1\_1 & 2K      & 0.2B  & 53\%  & 58\%      & 58\%   \\
conv1\_2 & 37K     & 3.7B  & 89\%  & 22\%      & 12\%   \\ \hline
conv2\_1 & 74K     & 1.8B  & 80\%  & 34\%      & 30\%   \\
conv2\_2 & 148K    & 3.7B  & 81\%  & 36\%      & 29\%   \\ \hline
conv3\_1 & 295K    & 1.8B  & 68\%  & 53\%      & 43\%   \\
conv3\_2 & 590K    & 3.7B  & 70\%  & 24\%      & 16\%   \\
conv3\_3 & 590K    & 3.7B  & 64\%  & 42\%      & 29\%   \\ \hline
conv4\_1 & 1M      & 1.8B  & 51\%  & 32\%      & 21\%   \\
conv4\_2 & 2M      & 3.7B  & 45\%  & 27\%      & 14\%   \\
conv4\_3 & 2M      & 3.7B  & 34\%  & 34\%      & 15\%   \\ \hline
conv5\_1 & 2M      & 925M  & 32\%  & 35\%      & 12\%   \\
conv5\_2 & 2M      & 925M  & 29\%  & 29\%      & 9\%    \\
conv5\_3 & 2M      & 925M  & 19\%  & 36\%      & 11\%   \\ \hline
fc6      & 103M    & 206M  & 38\%  & 4\%       & 1\%    \\
fc7      & 17M     & 34M   & 42\%  & 4\%       & 2\%    \\
fc8      & 4M      & 8M    & 100\% & 23\%      & 9\%    \\ \hline
total    & 138M    & 30.9B & 64\%  & \bf{7.5\%}  & \bf{21\%}  
\end{tabular}
\end{table}

\subsection{VGG-16 on ImageNet}
With promising results on AlexNet, we also looked at a larger, more recent network, VGG-16 \cite{Simonyan14c}, on the same ILSVRC-2012 dataset.  VGG-16 has far more convolutional layers but still only three fully-connected layers.  Following a similar methodology, we aggressively pruned both convolutional and fully-connected layers to realize a significant reduction in the number of weights, shown in Table~\ref{table:vggresults}. We used five iterations of pruning an retraining.

The VGG-16 results are, like those for AlexNet, very promising.  The network as a whole has been reduced to 7.5\% of its original size ($13\times$ smaller). In particular, note that the two largest fully-connected layers can each be pruned to less than 4\% of their original size.  
This reduction is critical for real time image processing, where there is little reuse of fully connected layers across images (unlike batch processing during training).

\section{Discussion}

\begin{figure}[t]
\centering
\scalebox{0.8}[0.8]{\includegraphics[scale=0.6]{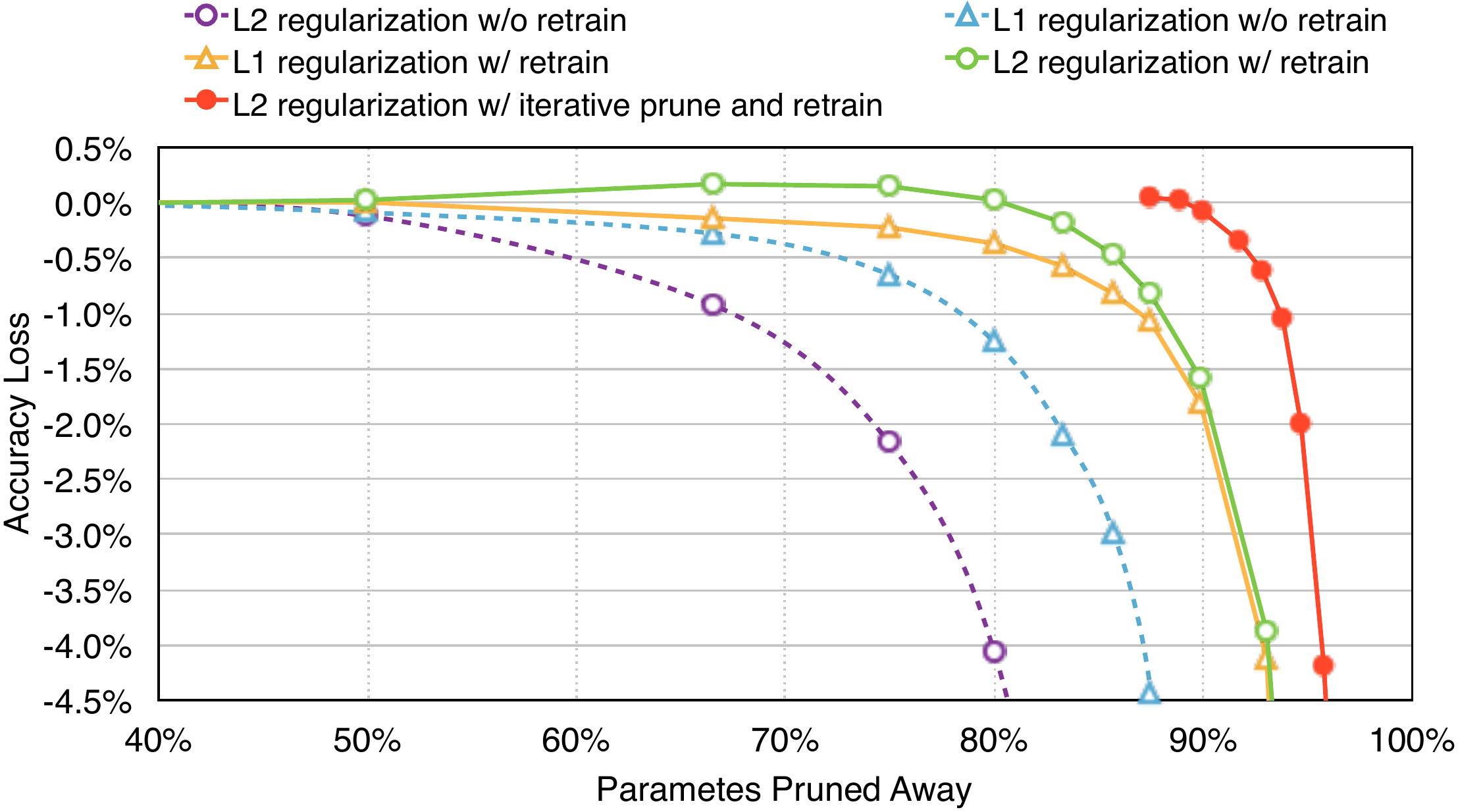}}
\caption{Trade-off curve for parameter reduction and loss in top-5 accuracy. L1 regularization performs better than L2 at learning the connections without retraining, while L2 regularization performs better than L1 at retraining. Iterative pruning gives the best result.}
\label{fig:acc_param}
\end{figure}

\begin{figure}[t]
\begin{minipage}[c]{0.4\textwidth}
\vspace{0pt}
\scalebox{0.6}[0.6]{\includegraphics[scale=0.55,angle=0]{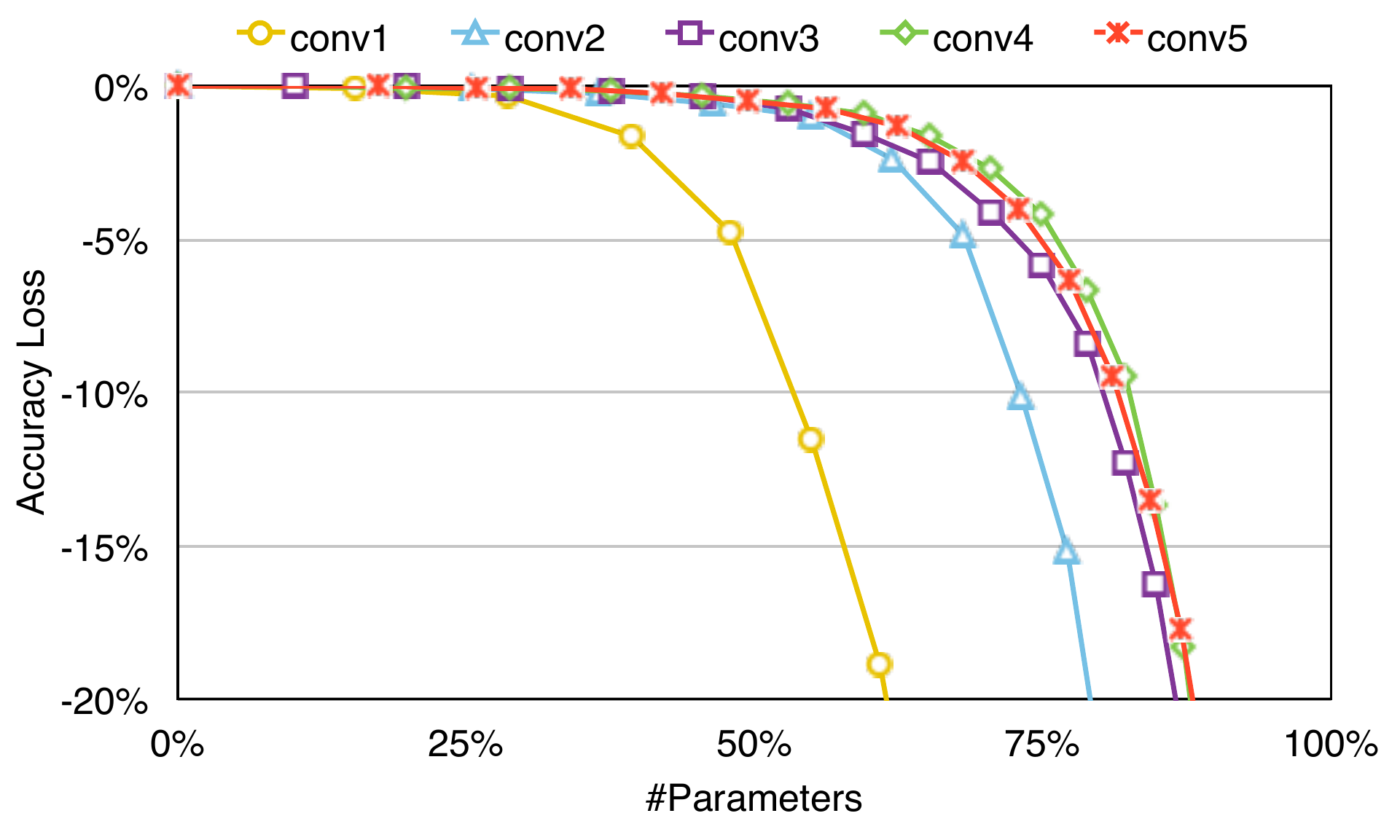}}
\end{minipage}
\hspace{0.1\textwidth}
\begin{minipage}[c]{0.4\textwidth}
\vspace{0pt}
\hspace{0pt}
\scalebox{0.6}[0.6]{\includegraphics[scale=0.55,angle=0]{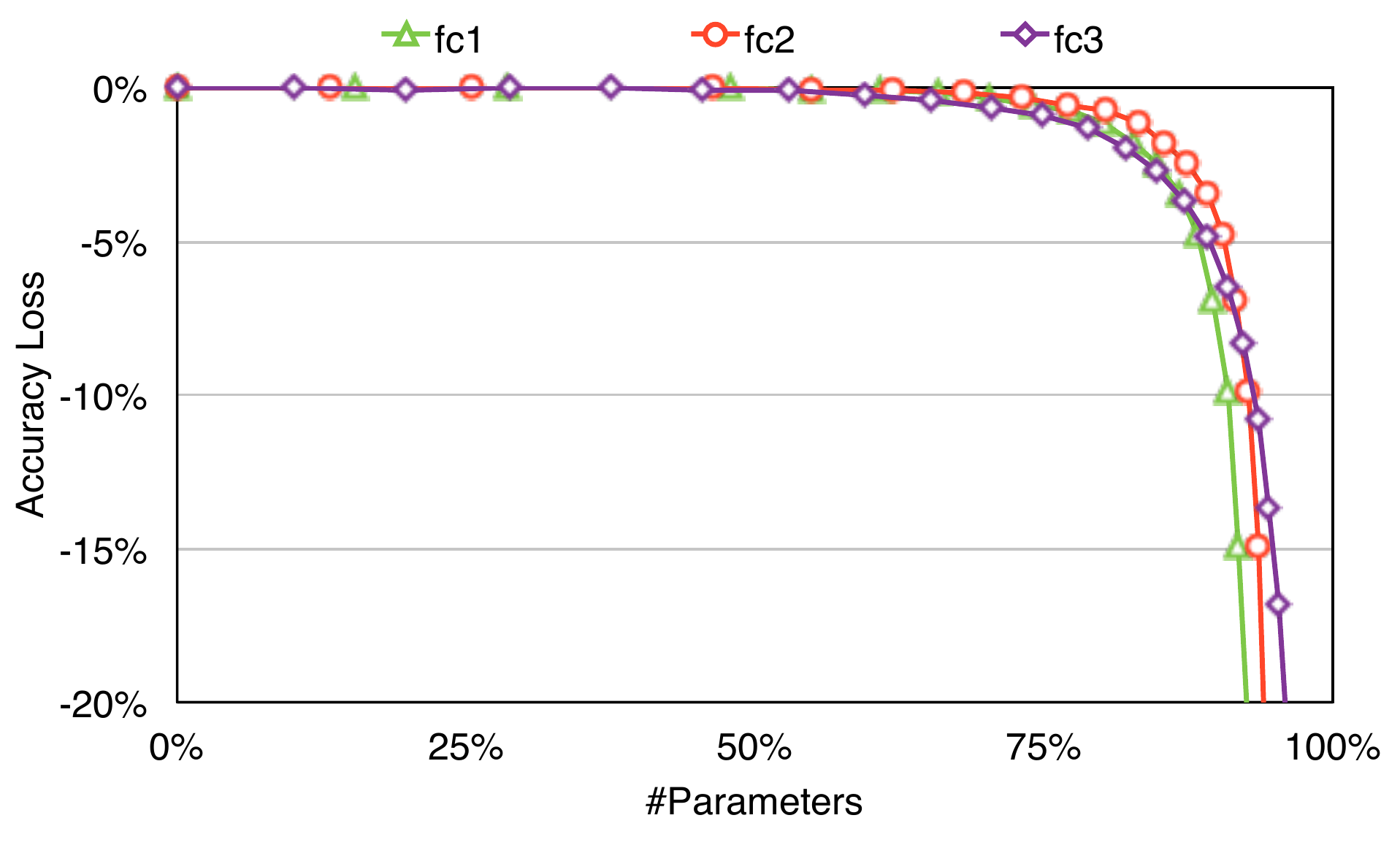}}
\end{minipage}
\caption{Pruning sensitivity for CONV layer (left) and FC layer (right) of AlexNet.}
\label{fig:sensitivity}
\end{figure}

The trade-off curve between accuracy and number of parameters is shown in Figure \ref{fig:acc_param}. The more parameters pruned away, the less the accuracy. 
We experimented with L1 and L2 regularization, with and without retraining, together with iterative pruning to give five trade off lines. 
Comparing solid and dashed lines, the importance of retraining is clear: without retraining, accuracy begins dropping much sooner --- with $1/3$ of the original connections, rather than with $1/10$ of the original connections. It's interesting to see that we have the ``free lunch'' of reducing $2\times$ the connections without losing accuracy even without retraining; while with retraining we are ably to reduce connections by $9\times$.

L1 regularization gives better accuracy than L2 directly after pruning (dotted blue and purple lines) since it pushes more parameters closer to zero.  However, comparing the yellow and green lines shows that L2 outperforms L1 after retraining, since there is no benefit to further pushing values towards zero.  One extension is to use L1 regularization for pruning and then L2 for retraining, but this did not beat simply using L2 for both phases.  Parameters from one mode do not adapt well to the other.

The biggest gain comes from iterative pruning (solid red line with solid circles). 
Here we take the pruned and retrained network (solid green line with circles) and prune and retrain it again.
The leftmost dot on this curve corresponds to the point on the green
line at 80\% ($5\times$ pruning) pruned to $8\times$. There's no accuracy loss at $9\times$.  Not until $10\times$ does the accuracy begin to drop sharply.

Two green points achieve slightly better accuracy than the original model. We believe this accuracy improvement is due to pruning finding the right capacity of the network and hence reducing overfitting.

Both CONV and FC layers can be pruned, but with different sensitivity. Figure \ref{fig:sensitivity} shows the sensitivity of each layer to network pruning.  
The figure shows how accuracy drops as parameters are pruned
on a layer-by-layer basis.
The CONV layers (on the left) are more sensitive to pruning than the fully connected layers (on the right). The first convolutional layer, which interacts with the input image directly, is most sensitive to pruning. We suspect this sensitivity is due to the input layer having only 3 channels and thus less redundancy than the other convolutional layers. We used the sensitivity results to find each layer's threshold: for example, the smallest threshold was applied to the most sensitive layer, which is the first convolutional layer.

Storing the pruned layers as sparse matrices has a storage overhead of only 15.6\%. Storing relative rather than absolute indices reduces the space taken by the FC layer indices to 5 bits.  Similarly, CONV layer indices can be represented with only 8 bits.

After pruning, the storage requirements of AlexNet and VGGNet are are small enough that all weights can be stored on chip, instead of off-chip DRAM which takes orders of magnitude more energy to access (Table \ref{energy}).  We are targeting our pruning method for fixed-function hardware specialized for sparse DNN, given the limitation of general purpose hardware on sparse computation.

\begin{table}[t]
\vspace{-4pt}
\centering
\caption{Comparison with other model reduction methods on AlexNet.
Data-free pruning \cite{srinivas2015data} saved only $1.5\times$ parameters with much loss of accuracy.
Deep Fried Convnets \cite{yang2014deep} worked on fully connected layers only and reduced the parameters by less than $4\times$.
\cite{collins2014memory} reduced the parameters by $4\times$ with inferior accuracy.  Naively cutting the layer size saves parameters but suffers from 4\% loss of accuracy. \cite{denton2014LinearStructure} exploited the linear structure of convnets and compressed each layer individually, where model compression on a single layer incurred 0.9\% accuracy penalty with biclustering + SVD.}
\begin{tabular}{lllll}
\multicolumn{1}{l|}{Network}              & Top-1 Error& \multicolumn{1}{l|}{Top-5 Error} & \multicolumn{1}{l|}{Parameters}  & \begin{tabular}[c]{@{}l@{}}Compression\\       Rate\end{tabular} \\ \hline
\multicolumn{1}{l|}{Baseline Caffemodel \cite{jia2014caffe}}  & 42.78\%     & \multicolumn{1}{l|}{19.73\%}         & \multicolumn{1}{l|}{61.0M}      & $1\times$ \\
\multicolumn{1}{l|}{Data-free pruning \cite{srinivas2015data}} & 44.40\%     & \multicolumn{1}{l|}{-}           & \multicolumn{1}{l|}{39.6M}      & $1.5\times$ \\
\multicolumn{1}{l|}{Fastfood-32-AD \cite{yang2014deep}}       & 41.93\%     & \multicolumn{1}{l|}{-}           & \multicolumn{1}{l|}{32.8M}      & $2\times$ \\
\multicolumn{1}{l|}{Fastfood-16-AD \cite{yang2014deep}}       & 42.90\%     & \multicolumn{1}{l|}{-}           & \multicolumn{1}{l|}{16.4M}      & $3.7\times$ \\
\multicolumn{1}{l|}{Collins \& Kohli \cite{collins2014memory}}& 44.40\%     & \multicolumn{1}{l|}{-}     & \multicolumn{1}{l|}{15.2M}            & $4\times$ \\
\multicolumn{1}{l|}{Naive Cut}            & 47.18\%          & \multicolumn{1}{l|}{23.23\%}          & \multicolumn{1}{l|}{13.8M}               & $4.4\times$ \\
\multicolumn{1}{l|}{SVD \cite{denton2014LinearStructure}}  & 44.02\%       & \multicolumn{1}{l|}{20.56\%}   & \multicolumn{1}{l|}{11.9M}      & $5\times$ \\
\multicolumn{1}{l|}{\bf{Network Pruning}}       & \bf{42.77}\%     & \multicolumn{1}{l|}{\bf{19.67\%}}     & \multicolumn{1}{l|}{\bf{6.7M}}           & $\bf{9\times}$     \\
\end{tabular}
\label{table:compare}
\end{table}

\vspace{-4pt}
\begin{figure}[t]
\begin{minipage}[c]{0.4\textwidth}

% \centering
\vspace{0pt}
\scalebox{0.6}[0.6]{\includegraphics[scale=0.35,angle=0]{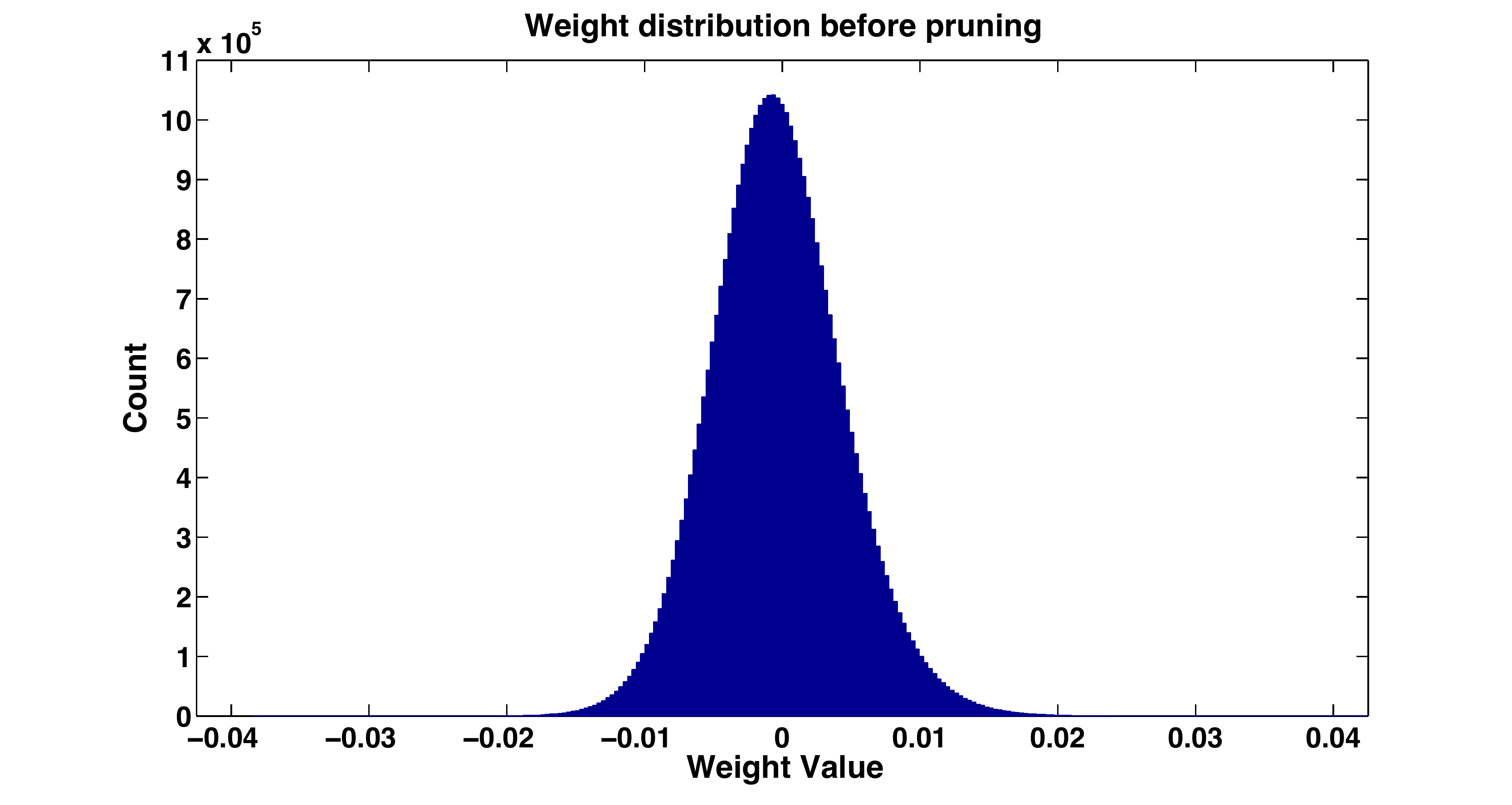}}
\end{minipage}
\hspace{25pt}
\begin{minipage}[c]{0.4\textwidth}
% \centering 

\scalebox{0.6}[0.6]{\includegraphics[scale=0.35,angle=0]{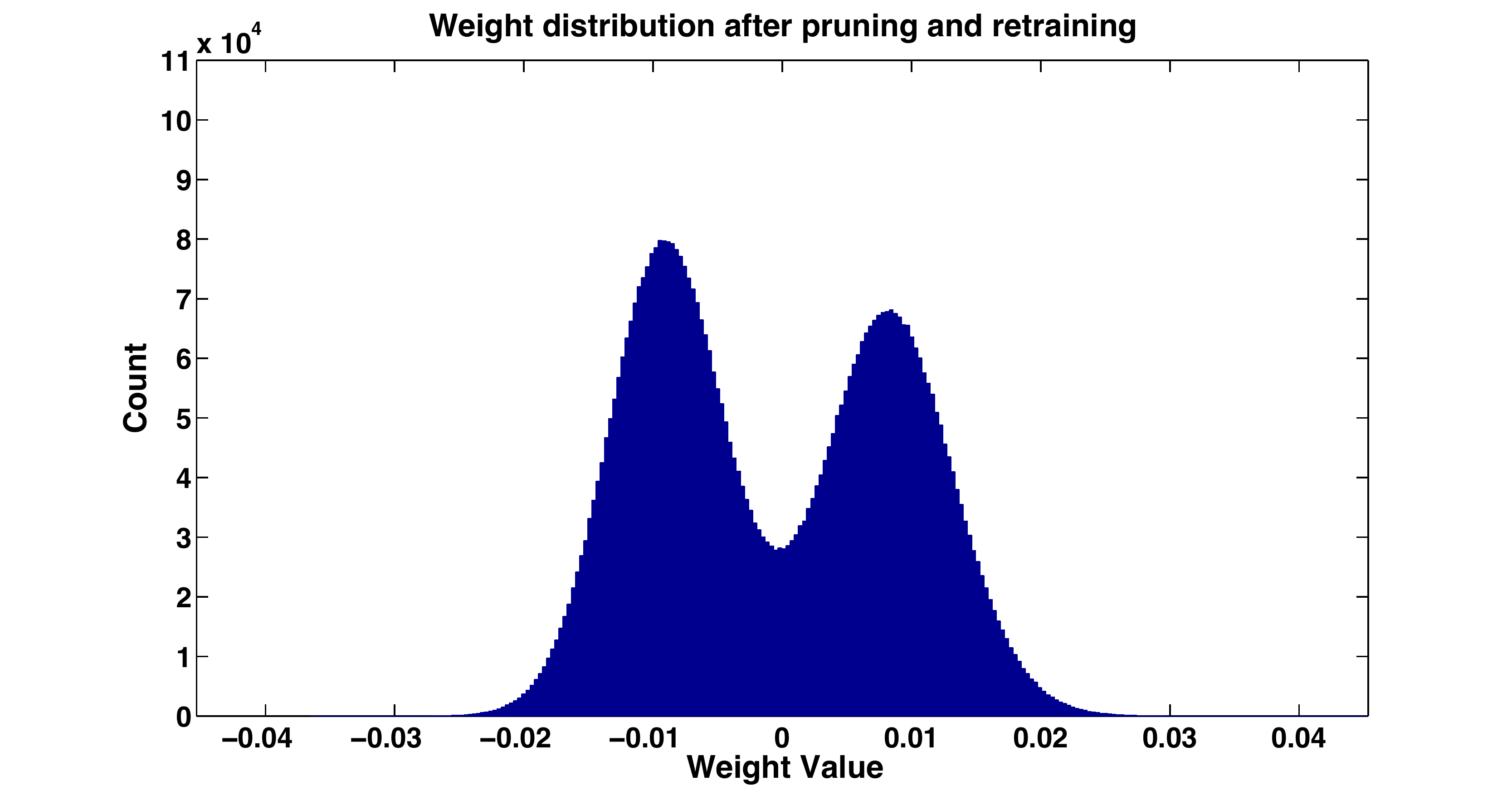}}
\end{minipage}
\caption{Weight distribution before and after parameter pruning. The right figure has $10\times$ smaller scale.}
\vspace{-5pt}
\label{fig:w1}
\end{figure}

Figure \ref{fig:w1} shows histograms of weight distribution before (left) and after (right) pruning. The weight is from the first fully connected layer of AlexNet. The two panels have different y-axis scales. 
The original distribution of weights is centered on zero with tails dropping off quickly. Almost all parameters are between $[-0.015, 0.015]$.
After pruning the large center region is removed. The network parameters adjust themselves during the retraining phase. The result is that the parameters form a bimodal distribution and become more spread across the x-axis, between $[-0.025, 0.025]$.

\section{Conclusion}
\vspace{-5pt}

We have presented a method to improve the energy efficiency and storage of neural networks without affecting accuracy by finding the right connections.
Our method, motivated in part by how learning works in the mammalian brain, operates by learning which connections are important, pruning the
unimportant connections, and then retraining the remaining sparse network.
We highlight our experiments on AlexNet and VGGNet on ImageNet, showing that both fully connected layer and convolutional layer can be pruned, reducing the number of connections by $9\times$ to $13\times$ without loss of accuracy.  This leads to smaller memory capacity and bandwidth requirements for real-time image processing, making it easier to be deployed on mobile systems.
\vspace{-5pt}

\small{
\bibliographystyle{unsrt}
\bibliography{nips2015.bib}
}

\end{document}